\documentclass[review]{elsarticle}
\usepackage{lineno,hyperref}
\modulolinenumbers[5]
\usepackage{booktabs}
\usepackage{multirow}
\usepackage{mathtools}
\journal{Journal of Applied Computing and Informatics}

\begin{document}

\begin{frontmatter}

\title{A novel framework of the fuzzy c-means distance’s problem based weighted distance}
\tnotetext[t1]{This document is a collaborative effort by Intelligent Systems Research Group Indonesia and Informatics Department Universitas Muhammadiyah Semarang.}






\author[1]{Andy Arief Setyawan\corref{cor1}%
	\fnref{fn1}}
\ead{andyariefsetyawan@gmail.com}
\author[2]{Ahmad Ilham\fnref{fn2}}
\ead{ahmadilham@unimus.ac.id}

\cortext[cor1]{Corresponding author}

\address[1]{Department of Information and Communication, Pemalang District Government, Pemalang, Indonesia}
\address[2]{Department of Informatics, Universitas Muhammadiyah Semarang, Semarang 50354, Indonesia}

\begin{abstract}

Clustering is one of the major roles in data mining that is widely application in pattern recognition
and image segmentation. Fuzzy C-means (FCM) is the most used clustering algorithm that proven
efficient, fast and easy to implement, however FCM uses the Euclidean distance that often leads to
clustering errors, especially when handling multidimensional and noisy data. In the last few years,
many distances metric have been propose by researchers to improve the performance of the FCM
algorithms, and the majority of researchers propose weighted distance. In this paper, we proposed
Canberra Weighted Distance to improved performance of the FCM algorithm. Experimental result
using the UCI data set show the proposed method is superior to the original method and other clustering methods.
\end{abstract}


\end{frontmatter}


\section{Introduction}

Cluster analysis or clustering is the process of partitioning a set of data objects into subset or clusters, where the objects in a cluster is  similar to one another and dissimilar to the objects on other clusters \cite{aggarwal2014data,han2011data,ilham2019}. Clustering algorithms appear as the formal tools for the computer-aided detection of the naturally occurring groups in a collection of objects or data set \cite{saha2015automated}. Clustering is the task for grouping a set of objects into a number of clusters by analyse similarity of some feature of object.

In general, clustering methods classified into four categories \cite{han2011data}:
\begin{enumerate}[(1)]
	\item \textit{Partition methods}: based on distance metric between object and centroid iteratively until convergence
	\item \textit{Hierarchical methods}: create hierarchical decomposition of given dataset
	\item \textit{Density-based methods}: find clusters of arbitrary shape, dense regions in the data space, separated by sparse regions
	\item \textit{Grid-based methods}: quantize the object space into a finite number of cells that form a grid structure.
\end{enumerate}

Hierarchical and partition methods is the most popular used clustering algorithm \cite{ferreira2016kernel}. Compared self-organizing map neural network, FCM, k-means and traditional hierarchical clustering on 2530 dataset, FCM proved superiority and stability for all case even addition by outlier and overlapping \cite{mingoti2006comparing}. Using UCI dataset, performance of the FCM algorithm better than k-means \cite{sivarathri2014experiments}. FCM also proving better performance over k-means for image segmentation process \cite{mirghasemi2016new}. Several recent studies are proving the stability of the FCM algorithm over other clustering methods.

FCM is one of the most used partition methods clustering algorithm because it is naturally characteristic. FCM algorithm introduce by Dunn \cite{dunn1973fuzzy} with Fuzzy ISODATA and developed later by Bezdek \cite{bezdek1984fcm}. Different with k-means where an object is belong to exactly one cluster, FCM allows an object belong to two or more clusters with a membership grade between zero and one \cite{wu2012advances}. Although FCM has performed well in cluster detection and proven superiority to applied in many applications, it has two interesting weakness to be improvement. The first is initialization of cluster center always set randomly, the efficiency of FCM highly depends on the initialization step, because the iterative process easily falls into a locally optimal solution \cite{benaichouche2013improved}. The second is original FCM is used to calculate the distance with Euclidean distance, however, it proved to be worse than the weighted distance \cite{bezdek1984fcm}, The FCM algorithm is based on the Euclidean distance metric where handle each feature of the dataset with the same proportion to determine the cluster of data point. However, some of the real-world datasets have features with different contributions in determining the cluster of data point.

For many pattern recognition problems such as classification, clustering, and retrieval problems, distance metric become an importance function to define similar and dissimilarity objects \cite{cha2007comprehensive}. By default, FCM algorithm used Euclidean distance to calculate the distance between data point and cluster centers. Another distance matric proven superiority to improve the FCM performance as Canberra distance \cite{jafar2012study,das2013pattern}, city block distance \cite{liao2013new}, minkowski distance \cite{soeleman2015automatic}. Result for the six different dataset characteristic, compared cosine, city block, chebychev and euclidean distance, canberra distance show its stability in handling various types of datasets \cite{charulatha2013comparative}. It is clear, that used another distance metric on FCM algorithm can improve clustering performance.

To extracting meaningful information from the real-world dataset that corrupted by noise, it is not enough used distance metric only. Feature weighted is the most used task to get more improvement in clustering application. Several statistical methods like entropy \cite{delgado2016environmental,su2010new,zhou2016fuzzy} proposed to calculate feature weighted. Mean variance and standard deviation \cite{su2010new,zhou2016fuzzy} used to measure similarity data point and feature weighted incorporate with distance metric of the FCM algorithm. By using feature weighting, FCM performance can be better, because each feature has a different contribution in determining its class.

There have been several proposed methods in recent years to deal with FCM weaknesses in determining distances metric, primarily by combining distance function and feature weighting. In 2004, Wang et al. \cite{wang2004improving} proposed Weighted Fuzzy C-means (WFCM), by combining city block distance and feature weighting from fuzziness entropy dataset. Experimental result on iris dataset show, error rate of the WFCM algorithm about 8/150, better than FCM with 16/150. By still using entropy to calculate feature weight in Suet al. \cite{su2010new}, proposed Entropy Weighting FCM (EWFCM). Experimental result on iris dataset, EWFCM can outperform WFCM with error rate 7/150.

In 2008, to get better feature weighting, Hung et al. \cite{hung2008bootstrapping} using bootstrapping approach and trying several fuzziness value on FCM algorithm, the best performance on iris dataset when using 10 fuzziness value with error rate 7/150, just same like EWFCM. Zhang et al. \cite{zhang2014interval},  in 2014 proposed fuzzy c-means clustering by genetically guided alternating optimization (GIWFCM), using genetic heuristic strategy has error rate 5/150 on iris dataset. Integrated improved feature weighting and mahalanobis distance, in 2014 Xing and Ha \cite{xing2014further} successful minimize error rate to 5/160 just same as GIWFCM, it is called Improved feature weighting fuzzy c-means (IFWFCM).

The success of clustering process on FCM algorithm strongly influenced on the selection of exact distance metric function and the weighting of features that match the dataset characteristics. This study uses experimental method, by comparing several distance metrics to found the best performance distance measurement if applied to FCM algorithm. In addition to distance measurements, the application of some feature weightings such as the mean value, entropy, standard deviation, variance and derivatives are also comparisons to obtain appropriate weighting methods on the FCM algorithm.

In this paper, we proposed Canberra Weighted Fuzzy C-means (CWFCM). Combining Canberra distance with feature weighted used fuzziness of variance to mean ratio (VMR)[28], CWFCM has outperform with other clustering methods. To find out the improved performance of the proposed method, this study uses five public datasets from UCI Machine Learning Repository, i.e. iris, wine, Wisconsin diagnostic breast cancer (WDBC), sonar and balance. To evaluated improvement of the proposed method, we used external validation that based on historical information of the dataset \cite{rendon2011internal} i.e. Rand Index RI, Purity P, Accuracy Rate AR and Error Rate ER.

The rest of the paper organized in five sections. Section 2 is the basic of theory where described foundation of the FCM algorithm, various distance metric currently used in FCM and feature weight learning. Section 3 is the proposed algorithm CWFCM. Experiment for this research descripted in section 4, where contain three sub sections i.e. dataset currently used, evaluation methods, experimental result and statistical analysis that described in tabulated. Finally, section 5 is discuses about conclusion and future work of this paper.

\section{Material and algorithms}
\subsection{fuzzy c-means algorithm}

FCM algorithm partitions a set of \textit{j}-dimensional dataset 
$X=\left \{ x_1, x_2,..., x_n \right \}$, $\left ( 1\leq i\leq n 
\right )$ into \textit{c} clusters, every cluster is fuzzy set with 
membership degree $\mu_{ik}\left ( \in \left \lfloor 0, 1 \right 
\rfloor \right )$ indicating how much the sample $x_i$ belong to 
the cluster centers $v_jk$.  FCM aims determine cluster centers 
$v_kj \left ( 1\leq k\leq c \right )$, $\left ( 1\leq j\leq m 
\right )$ where \textit{m} i.e. dimension of the data set and the 
fuzzy partition matrix $\mu_{ik} \left ( 1\leq k\leq c \right )$ 
where \textit{n} i.e. sum of the data point $\left ( 1\leq k\leq c 
\right )$ by minimizing objective function as Equation \eqref{eq1}.
\begin{equation}
	P_{t}=\sum_{i=1}^{n}\sum_{k=1}^{c}\mu_{ik}^{d} d_{ij}^{2}
	\label{eq1}
\end{equation}
Where \textit{n} is sum of data point, \textit{c} is clusters, \textit{z} is fuzziness exponent, $\mu$ is the fuzzy partition matrix that calculate with Equation \eqref{eq2}.
\begin{equation}
	\mu_{ik}=\frac{\left ( d_{ij}^{2} \right )^\frac{-1}{z-1}}{\sum_{k=1}^{c}\left ( d_{ij}^{2} \right )^\frac{-1}{z-1}}
	\label{eq2}
\end{equation}
$d_{ij}^{2}$ is the distance between data point and cluster center, by default FCM using euclidean distance that calculate in Equation \eqref{eq3}.
\begin{equation}
	d_{ij}^{2}=\sum_{ij}^{m}\left | x_{ij} - v_{kj}\right |^2
	\label{eq3}
\end{equation}
Where $x_{ij}$ is the data point, and $v_{kj}$ is the cluster centers where compute as:
\begin{equation}
	v_{kj}=\frac{\sum_{i=1}^{n} \mu_{ik}^{m} x_{ij}}{\sum_{i=1}^{n} \mu_{ik}^{m}}
	\label{eq4}
\end{equation}

Descripted on Figure 1, FCM algorithm consists of several stages. Stage 1 is the input of the data set, where is the data point, j aim data dimension or feature space. Stage 2 – 3 is the initialize stages, fuzziness exponent m initialized with 2, in stages 3 fuzzy partition matrix $\mu_{ik}$ is assign with random values between 0 and 1. Stage 4 is compute cluster centre $v_{kj}$ with Equation \eqref{eq4}. Stage 5 calculate the distance using euclidean distance with Equation \eqref{eq3}. Stage 6 compute objective function P(t) with equation 2.1 and stage 7 updating fuzzy partition matrix $\mu\left ( i,k \right )$ with Equation \eqref{eq2}, do iteratively stage 4 – 7 until stopping criterion met. Table 1 is pseudocode for FCM algorithm that descripted all stage algorithmically.

\begin{figure}[h!]
	\begin{center}
		\includegraphics[width=11.0cm]{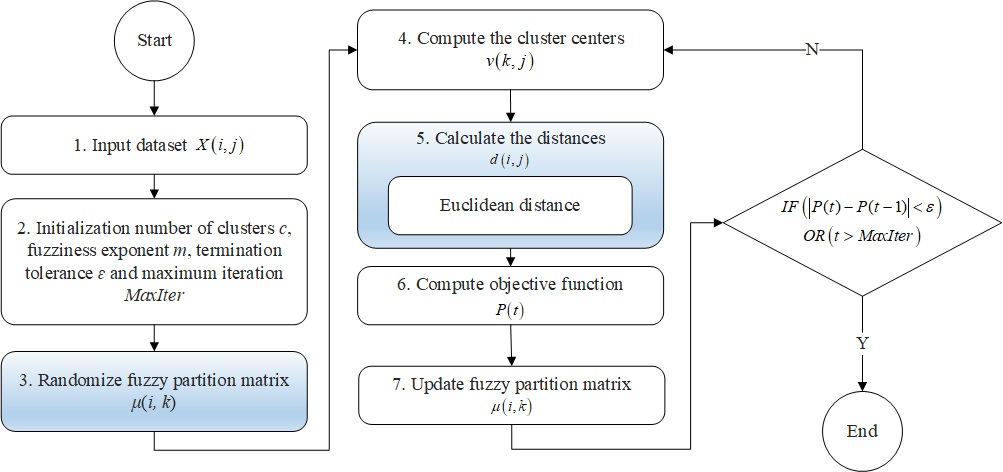}
		\caption{Flow diagram of the fuzzy c-means algorithm}
		\label{fig1}
	\end{center}
\end{figure}

\begin{table}[h!]
	\caption{Pseudocode of the fuzzy c-means}
	\begin{center}
		\begin{tabular}{@{}lll@{}}
		\toprule
		\multicolumn{3}{l}{\textbf{Algorithm 1 :} Fuzzy c-means} \\ \midrule
		\multicolumn{3}{l}{\begin{tabular}[c]{@{}l@{}}\textbf{Input:} Datasets $x\left ( i,j \right )$ \\ \textbf{Output:} terminal fuzzy partition matrix $ \mu \left ( i,k \right )$\\ \textbf{Begin}\end{tabular}} \\
		\midrule
		1 & \multicolumn{2}{l}{Input dataset $x\left ( i,j \right ) $} \\
		2 & \multicolumn{2}{l}{\begin{tabular}[c]{@{}l@{}}Initialization number of clusters \textit{c}, fuzziness exponent \textit{m}, termination tolerance $ \varepsilon $ \\ and maximum iteration \textit{MaxIter};\end{tabular}} \\
		3 & \multicolumn{2}{l}{Initialization fuzzy partition matrix $ \mu \left ( i,k \right )$ randomize;} \\
		4 & \multicolumn{2}{l}{\textbf{Repeat}} \\
		5 &  & Compute the cluster centers using Eq. (4); \\
		6 &  & Calculate distance using Eq. (3); \\
		7 &  & Compute objective function using Eq. (1); \\
		&  & Update fuzzy partition matrix using Eq. (2); \\
		 & \multicolumn{2}{l}{\textbf{Until} $\left ( \left | P\left ( t \right )-P\left ( t-1 \right ) \right | or \left ( t>MaxIter \right ) \right )$}  \\
		\multicolumn{3}{l}{\textbf{End}} \\ \bottomrule
	\end{tabular}
	\end{center}
\end{table}

\subsection{Distance matric}

In the last five years, improving distance metric in FCM algorithm become interesting topic in the field of data mining. By default, FCM algorithm used Euclidean distance, another Minkowski family $\left ( L_{p} \right )$ distance \cite{benaichouche2013improved} that usually used in FCM algorithm is city block or Manhattan distance \cite{mingoti2006comparing,xing2014further} this distance metric defines as Equation \eqref{eq5}.
\begin{equation}
d_{ij}=\sum_{j=1}^{m}\left | x_{ij}-v_{kj} \right |
\label{eq5}
\end{equation}

Minkowski distance also used to improve FCM performance for segmented moving object \cite{das2013pattern}, it is defined as Equation \eqref{eq6}.
\begin{equation}
d_{ij}=\sum_{j=1}^{m}\sqrt[p]{\left | x_{ij}-v_{kj} \right |^{p}}
\label{eq6}
\end{equation}
where \text{p} is the positive integer, when \text{p}=1 it became city block distance, when \textit{p}=2 it became Euclidean distance.

In the $L_{1}$ family distance, more precisely absolute difference, the most used distance in this family on FCM algorithm is Canberra distance \cite{bai2011mean}. This distance defines as Equation \eqref{eq7}.
\begin{equation}
d_{ij}=\sum_{j=1}^{m}\frac{\left | x_{ij}-v{kj} \right |}{\left | x_{ij}\right |+\left | v_{kj}\right |}
\label{eq7}
\end{equation}

Moreover, another distance metric usually used in FCM algorithm is mahalanobis distance, calculate by Equation \eqref{eq8}.
\begin{equation}
d_{ij}=\left ( x_{ij}-v_{kj} \right )^T A \left( x_{ij}-v_{kj}\right)
\label{eq8}
\end{equation}
where $\left (x_ij-v_kj \right )^T$ is the transpose matrix of $ \left( x_ij-v_kj \right ) $ and \textit{A} is positively semi-definite, $A \geq 0.1$ \cite{rendon2011internal}. When \textit{A}=1, gives euclidean distance, if \textit{A} is restrict to be diagonal, it is given different weight.

\subsection{Feature weighted}

In real world datasets, especially for the high dimensional datasets, every feature may have different degrees of relevance \cite{hu2016improved}. Feature weighting is very important task that must be have more attention during clustering process. There are several methods to decide feature weight on the datasets, that is filter methods, wrapper methods and embedded methods \cite{lance1966computer}. In this study, we using filter methods to get weight of the relevance feature from every dataset. Statistical functions to obtain suitable feature weight for each feature is variance to mean ratio \cite{wang2004improving} that calculate as:
\begin{equation}
VMR=s^{2}/x^{'}
\label{eq9}
\end{equation}
where $s^{2}$ is the sample variance of datasets and $x^{'}$ is mean for each feature, it is define in Eq. (\eqref{eq10}).
\begin{equation}
s^{2}=\frac{1}{n-1}\sum_{i=1}^{n}\left ( x_{i} - x^{'} \right ), x^{'}=\frac{\sum_{i=1}^{n}}{x_{i}}
\label{eq10}
\end{equation}
where, $x_{i}$ is the \textit{i}-th data point and \textit{n} is the sum of data point.

\subsection{Proposed method}
The proposed CWFCM method is integration of Canberra distance and feature weighted using fuzziness of \textit{VMR} from each feature to determine similarity and dissimilarity data point and centroids. This method is effective to deal with the second problem of FCM algorithm i.e. distances metric.

To deal with the first problem of FCM algorithm i.e. random initialize that can make inefficient result, we used fuzziness of sum square root from each feature to initialize partition matrix. Partition matrix $\mu_{ik}$ is defined as:
\begin{equation}
\mu_{i k}=\left\{\begin{array}{ll}{0} & {S F_{i}=a} \\ {\left(S F_{i}-a\right) /(a-b)} & {a<S F_{i}<b} \\ {1} & {S F_{i}=b}\end{array}\right.
	\label{eq11}
\end{equation}
where, \textit{a} is the smallest value and b is the largest value of SF set. SF is the sum of square root from values for each feature data point that calculate as:
\begin{equation}
SF_{i}=\sum_{j=1}^{m}\sqrt{{x_{ij}}^{2}}
\label{eq12}
\end{equation}
were, \textit{m} is sum of features for each \textit{x} data point.

\begin{figure}[h!]
	\begin{center}
		\includegraphics[width=11.0cm]{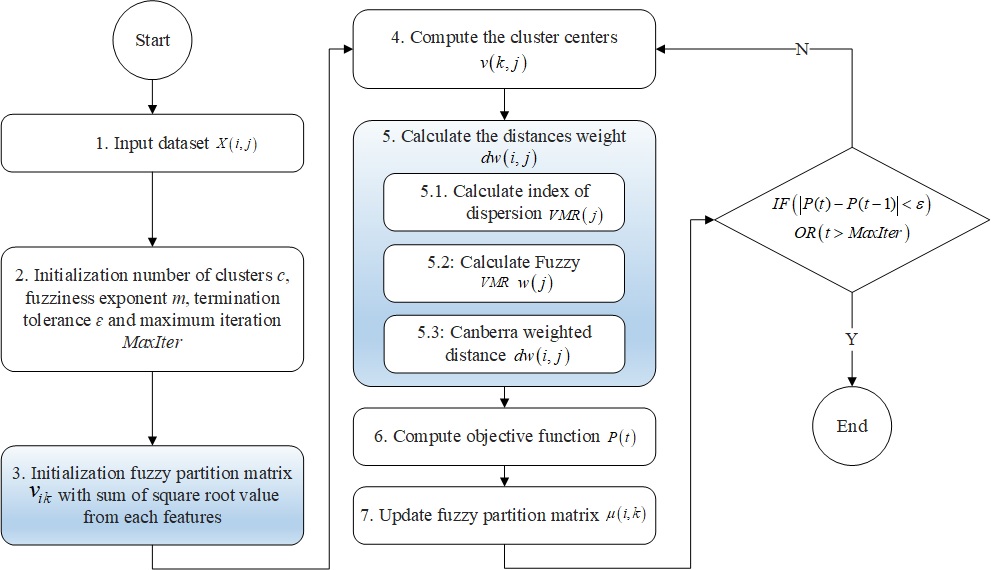}
		\caption{Flow diagram of the proposed algorithm}
		\label{fig2}
	\end{center}
\end{figure}

\begin{table}[h!]
	\centering
	\caption{Pseudocode of the proposed method}
	\begin{tabular}{@{}lll@{}}
		\toprule
		\multicolumn{3}{l}{\textbf{Algorithm 1 :} Fuzzy c-means} \\ \midrule
		\multicolumn{3}{l}{\begin{tabular}[c]{@{}l@{}}\textbf{Input:} Datasets $x\left ( i,j \right )$ \\ \textbf{Output:} terminal fuzzy partition matrix $ \mu \left ( i,k \right )$\\ \textbf{Begin}\end{tabular}} \\
		\midrule
		1 & \multicolumn{2}{l}{Input dataset $x\left ( i,j \right ) $} \\
		2 & \multicolumn{2}{l}{\begin{tabular}[c]{@{}l@{}}Initialization number of clusters \textit{c}, fuzziness exponent \textit{m}, termination tolerance $ \varepsilon $ \\ and maximum iteration \textit{MaxIter};\end{tabular}} \\
		3 & \multicolumn{2}{l}{Initialization fuzzy partition matrix $ \mu \left ( i,k \right )$ randomize;} \\
		4 & \multicolumn{2}{l}{\textbf{Repeat}} \\
		5 &  & Compute the cluster centers using Eq. (4); \\
		6 &  & Calculate Canberra weighted distance using Eq. (13); \\
		7 &  & Compute objective function using Eq. (1); \\
		&  & Update fuzzy partition matrix using Eq. (2); \\
		& \multicolumn{2}{l}{\textbf{Until} $\left ( \left | P\left ( t \right )-P\left ( t-1 \right ) \right | or \left ( t>MaxIter \right ) \right )$}  \\
		\multicolumn{3}{l}{\textbf{End}} \\ \bottomrule
	\end{tabular}
\end{table}

Fig. \eqref{fig2} shows a flowchart of the proposed algorithm. The stage of the proposed method same is FCM algorithm stage except for stage of 3 and 5. Stage 3 is defining initialize partition matrix, defined in Eq. \eqref{eq11}. Stage 5 calculate distance described in three sub stage i.e. calculate VMR or index of dispersion, calculate fuzziness VMR and compute metric using Canberra distance and feature weighted from fuzziness VMR. Canberra weighted distance is defined in Eq. (\eqref{eq13}).
\begin{equation}
	d_{ij}^{w}=d_{ij}*w_{j}
	\label{eq13}
\end{equation}
where, $d_{ij}$ is the Canberra distance that define by Equation \eqref{eq7}. For better validity result, feature weight distribution appropriate in interval 0 to 1 \cite{kang2016weight}. Feature weight $w_{j}$ is fuzzy set that define as:
\begin{equation}
w_{j}=\left\{\begin{array}{ll}{0} & {V M R_{j}=a} \\ {\left(V M R_{j}-a\right) /(a-b)} & {a<V M R_{j}<b} \\ {1} & {V M R_{j}=b}\end{array}\right.
\label{eq14}
\end{equation}

With a being the smallest value of VMR from each feature and b is the largest value. The weighting determination using Equation \eqref{eq14} reduces the number of features, i.e. for features with the smallest \textit{VMR}, so that, the feature with the smallest \textit{VMR} value has a zero weight and features with the largest \textit{VMR} value has a weight of one. 

Table 2, shows a description of the pseudocode algorithmically for all stages in a flowchart of the proposed method in Fig. \eqref{fig2}

\section{Experimental results and analysis }
\subsection{Datasets}
Many studies by distance metric learning on the FCM algorithm are using public datasets. In this study, public datasets also used to test the proposed method. Public datasets mostly used to this study were available in UCI machine learning repository. Table \eqref{tab:table3} shows description of the datasets, from the left column is name of the datasets, number of instances, number of attributes, number of class and data distribution for each class. From the UCI machine learning repository, the most widely used datasets are selected and have varying dimensions and number of classes, so that can represent all conditions in testing the stability of the proposed method.
\begin{table}[h!]
	\caption{Dataset description}
	\begin{center}
	\begin{tabular}{lllll}
		\toprule
		Datasets	& Instances		& Attrb.	& Class	& Distribution	\\
		\midrule
		Iris & 150  & 4	& 3 	& 50, 50, 50     \\
		Sonar	& 208 & 60	& 2	& 97, 111      \\
		WDBC     & 569	& 30	& 2	& 357, 212  \\
		WBC	& 699 & 9  & 2  & 458, 241  \\  
		Wholesale Customers	& 582  & 7  & 2  & 298, 142  \\ 
		Abalone	& 4177	& 8	& 3	& 1528, 1307, 1342  \\ 
		Tae		& 100	& 5	& 3	& 22, 33, 45  \\  
		Yeast	& 1484	& 8	& 10	& \begin{tabular}[c]{@{}l@{}}244, 429, 463, 44, 35, \\ 51, 163, 30, 20, 5\end{tabular}  \\  
		Bufa	& 245	& 6	& 2	& 145, 200, 345 \\
		\bottomrule
	\end{tabular}
	\end{center}
	\label{tab:table3}
\end{table}

\subsection{Evaluation}
In general, there are two types of evaluation for the clustering algorithm, i.e. external validation by utilizing information on the historical data and internal validation by simply utilizing the intrinsic information of the data itself \cite{xing2014further}. To know the proposed performance improvement, this research using external validation method, that is rand index \cite{de2016survey}, purity \cite{chandrashekar2014survey}, and error rate ER.

Rand index (RI) has a range of values between zero and one, the highest value indicates better performance, this evaluation described as:
\begin{equation}
R I=\frac{a+d}{a+b+c+d}
\label{eq15}
\end{equation}

For a set of n object $X=\left\{x_{1}, x_{2}, \ldots, x_{n}\right\}$, if $Y=\left\{y_{1}, y_{2}, \ldots, y_{k}\right\}$ is result of the clustering method and $Z=\left\{z_{1}, z_{2}, \ldots, z_{k}\right\}$ is the actual class. The \textit{a} is number of pairs \textit{X} were in the same subset \textit{Y} and same subset in \textit{Z}, \textit{b} = number of pairs \textit{X} were in the different subset \textit{Y} and different subset in \textit{Z}. The \textit{c} is a number of pairs \textit{X} were in the same subset \textit{Y} and different subset in \textit{Z}. In addition, \textit{d} is a number of pairs \textit{X} were in the different subset \textit{Y} and same subset in \textit{Z}.

Purity P focuses on the frequency of the most common category into each cluster \cite{chandrashekar2014survey} and define as:
\begin{equation}
P=\frac{1}{n} \sum_{j=1}^{c} d_{j}
\label{eq16}
\end{equation}
where \textit{n} is sum of the data point, \textit{c} is sum of clusters and $d_{j}$ is sum of data point that classified exactly on the \textit{j} cluster. 

Error rate ER is defined as:
\begin{equation}
E R=\left(1-\frac{c p}{n}\right) * 100
\label{eq17}
\end{equation}
where, \textit{cp} is the wrong classified data point and \textit{n} is the sum of data point.

To evaluated efficiency for proposed method, this study using iteration that explained how many time algorithms looping until stopping criterion meet.

\subsection{Experiment results}
The computer specifications used to simulate the performance of the proposed algorithm descripted on Table \eqref{tab:table4}. The proposed algorithm is implement using basic language on Ms. Visual Studio 2012. The add noise function on RapidMiner 7.5 is used to generate additional noise on the datasets.
\begin{table}[h!]
	\caption{Computer specs}
	\begin{center}
	\begin{tabular}{@{}ll@{}}
		\toprule
		Items & Specs \\ \midrule
		Processor & Intel Core i5-5200U 2.7 Ghz \\
		Memory & 6 GB \\
		Hard Disk & 1 TB \\
		Operation System & Windows 10 Professional 64 bit \\
		Software apps & Ms. Visual Studio 2012 \\
		& RapidMiner Studio 7.5 \\
		& XLSTAT \\ \bottomrule
	\end{tabular}
	\label{tab:table4}
	\end{center}
\end{table}

\begin{table}[h!]
	\caption{Comparison error rates and misclassification with other method}
	\begin{center}
		\begin{tabular}{@{}llll@{}}
		\toprule
		\multirow{2}{*}{Authors} & Feature weight & \multirow{2}{*}{ER} & \multirow{2}{*}{MC} \\ \cmidrule(lr){2-2}
		& SL, SW,  PL, PW &  &  \\ \cmidrule(r){1-4} 
		Bezdek 1981 \cite{bezdek1984fcm} & - & 10.667 & 16 \\
		Wang et al. 2004 \cite{wang2004improving} & 0.0001, 0.0002, 1, 0.164 & 5.333 & 8 \\
		Hung et al. 2008 \cite{hung2008bootstrapping} & 0.102, 0.1022, 0.3377, 0.458 & 4.667 & 7 \\
		Zhang et al. 2014 \cite{zhang2014interval} & 0.0571, 0.1058, 0.4505, 0.3867 & 3.333 & 5 \\
		Xing et al. 2014 \cite{xing2014further} & 0.1194, 0.1134, 0.4346, 0.3327 & 3.333 & 5 \\
		Proposed method & 0.0728, 0, 1, 0.5534 & \textbf{2.667} & \textbf{4} \\ \bottomrule
	\end{tabular}
	\label{tab:table5}
	\end{center}
\end{table}

Table \eqref{tab:table5} showing comparison of proposed method with other FCM methods on the iris dataset. The second column is feature weight for each algorithm except Bezdek \cite{bezdek1984fcm} or native methods were not present by feature weighted. The third column is percentage of error rates for each algorithm and the fourth column is data point that misclassification for each algorithm.

Still using iris dataset, Table \eqref{eq6} showing comparison several distance matric with three feature weighted on FCM algorithm. The first column descripted three kind of feature weighted, second column is additional noise from 0 to 30 for each feature weighted and the third column until eighth is distance metric i.e. Euclidean (Eucl.), City Block (CB), Canberra (Canb.), Minkowsky (Mink.) and Mahalanobis (Maha.) distance. The first row for every grouping is iteration that consume by the methods, for left side by using random initialize and the right one is using proposed initialize. The second row for every grouping is percentage of error rates for the methods.

\begin{table}[h!]
	\begin{center}
		\caption{Comparison iteration and error rates on iris dataset}
		\includegraphics[width=9.0cm]{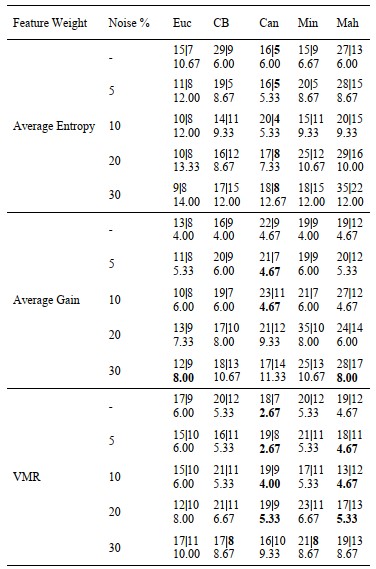}
		\label{tab:table6}
	\end{center}
\end{table}

As you can see in Table \eqref{tab:table6}, the proposed initialization is always superior to random initialization and smallest iteration on every dataset is when using average entropy and Canberra distance with 5, 5, 4, 8 and 8. The best performance produces error rates for all dataset is when using VMR and Canberra distance, except for 30$\%$ additional noise, average gain with Euclidean and Mahalanobis distance difference of 1.33, better than proposed method with 8.00$\% $ error rates.

\begin{table}[h!]
	\begin{center}
		\caption{Comparison iteration and error rates on WDBC dataset}
		\includegraphics[width=9.0cm]{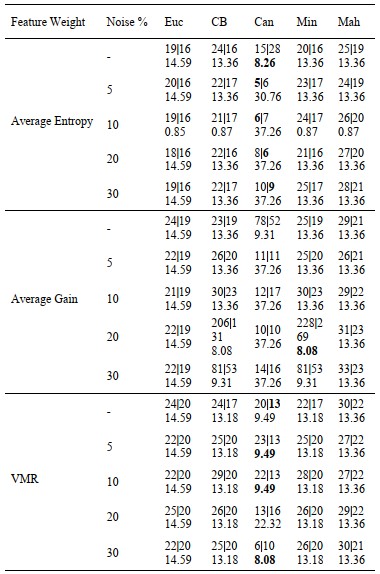}
		\label{tab:table7}
	\end{center}
\end{table}

With using VMR feature weighted as descripted on Table \eqref{tab:table6} and Table \eqref{tab:table7}, canberra distance shows the best performance on each dataset. The proposed algorithm shows its stability to handle both datasets even with several additional noise. The proposed method also proven efficiency with iteration between 7 and 10 for iris and 13 until 16 for WDBC. This proves that the use of feature weights using \textit{VMR} on canberra distance is very influential on the success of the clustering process. 
\begin{table}[h!]
	\caption{Comparison of average iteration}
	\begin{center}
		\begin{tabular}{@{}lll@{}}
		\toprule
		\multirow{2}{*}{Datasets} & \multicolumn{2}{l}{Initialize} \\ \cmidrule(l){2-3} 
		& Random & Proposed \\ \midrule
		Iris & 18 & 10 \\
		WDBC & 28 & 23 \\ \midrule
		Total & 46 & 33 \\ \bottomrule
	\end{tabular}
	\end{center}
	\label{tab:table8}
\end{table}

To describe superiority of the proposed initialization method, Table \eqref{tab:table8} is a comparison of the results of the average number of iterations obtained when using VMR feature weighting and canberra distance as outlined in Table 6 and Table \eqref{tab:table7}. Can be seen that, the average initialization proposed in this study has fewer iterations than all random models, this proves from the side of computational time, the proposed method is more efficient than the previous method.

Just like IFWFCM, CWFCM improvised FCM stages by changing the initialization process, features weighted and changing euclidean distances using canberra distance. To test the stability of CWFCM, Table \eqref{tab:table9}, shows comparison of the proposed method with the previous method in the terms of computational cost. This study uses average of iterations and time in seconds that's needed until the stopping criterion of the algorithm are met or the partition is declared optimal. Each method is run 10 times in each dataset, then the average value of the iteration results and the computational times in seconds is taken to obtain an objective comparison of the scores.

\begin{table}[h!]
	\begin{center}
		\caption{Comparison of iteration and times on 9 datasets}
		\includegraphics[width=11.0cm]{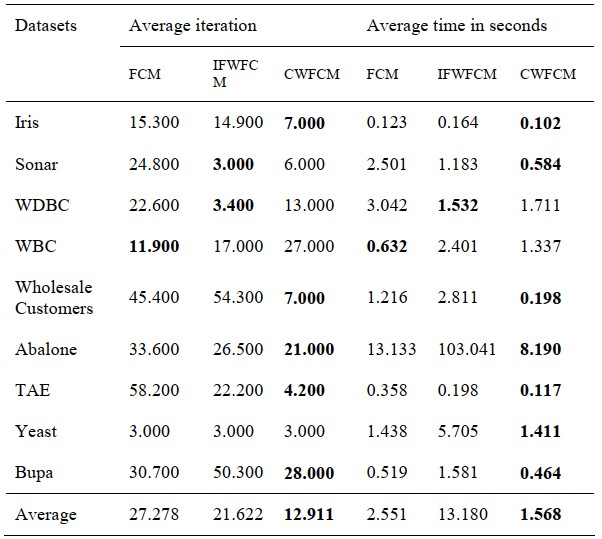}
		\label{tab:table9}
	\end{center}
\end{table}

As can be seen in Table \eqref{tab:table9}, the number of iterations is not always directly proportional to the time needed to reach the optimal partition. The most significant difference is seen in the abalone dataset, the average number of FCM iterations is 33.6 while the IFWFCM iteration average is smaller at 26.5, but FCM takes less time i.e. 13.133 seconds while IFWFCM takes 103,041 seconds, more than 1 minute to achieve optimal conditions, in this research, this is the longest time ever. In the same dataset CWFCM only requires 21 iterations with 8.19 seconds needed.

Judging from the time needed to meet optimum conditions, CWFCM is only longer on the WDBC and WBC datasets. In the WDBC dataset the fastest time is IFWFCM with 1.532 seconds, 0.179 seconds faster than CWFCM. As for the WBC dataset, the fastest method is FCM with 0.632 seconds, 0.705 seconds faster than CWFCM with 1,337 secondson average CWFCM outperforms other methods with 1.568 second, while FCM 2.551 seconds and IFWFCM takes an average of 13.180 seconds to reach optimal conditions.As you can see from the results of the table explanation, that CWFCM can be stated to have more efficient computational costs.

\begin{table}[h!]
	\begin{center}
		\caption{Comparison of purity, rand index and error rates on 9 datasets}
		
		\includegraphics[width=11.0cm]{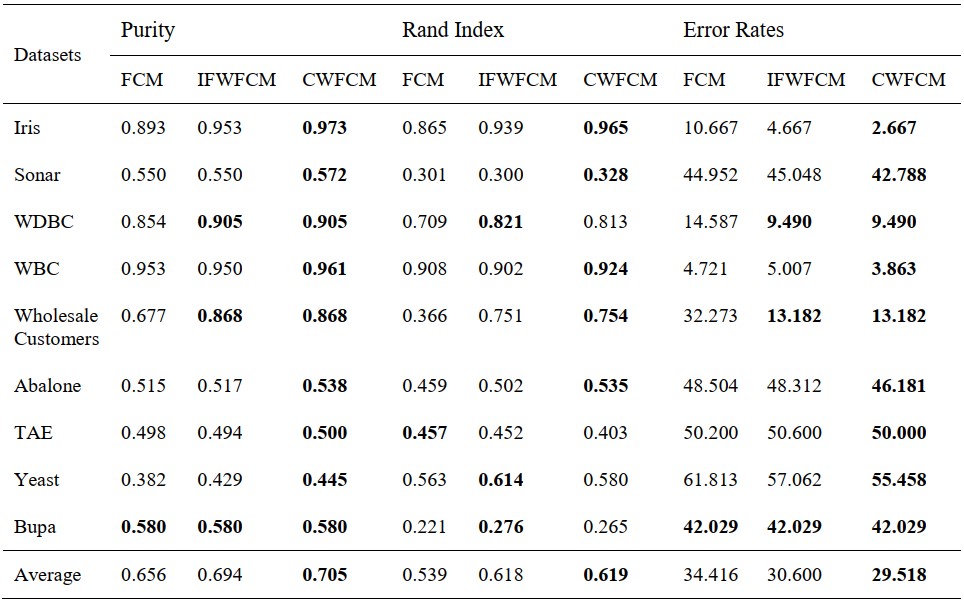}
		\label{tab:table10}
	\end{center}
\end{table}

The computational costs required by each method described in Table 9 cannot be stated as the only measure that a method with minimum costs is at least superior to a method that has more computational costs. To be more convincing about superiority of the proposed method, Table 10 presents the results of a comparison of methods using various external validation. In the term of Purity, CWFCM has the best performance in all datasets, even though IFWFCM has the same results as CWFCM on 3 datasets namely WDBC, Wholesale Customers and Bupa, as well as FCM which has the same results on the Bupa dataset. On average CWFCM obtained a purity value of 0.705, superior to 0.011 from IFWFCM and outperformed 0.049 from FCM.

In addition, to further prove the superiority of the proposed method, although it is not as perfect as using Purity, with using Rand Index, from 9 datasets, CWFCM excels at 5 datasets, while IFWFCM excels in 3 datasets i.e. WDBC, Yeast and Bupa, while FCM excels in only one dataset i.e. TAE. However, on average even though it has a difference of only 0.001 from IFWFCM, CWFCM still excels with an average rand index of 0.619, and superior to FCM with a difference of 0.080.

Finally, to determine the level of cluster error, this study uses the error rates. Error rates are displayed in the rightmost column group, error rates can be used to find out the number of instances that can't be correctly classified or as mentioned misclassification (MC) on Table 5.Number of MC are calculated by equation:error rates * total number of instant / 100, for example in the iris dataset the number of instant that is not properly classified is 4 instant, obtained from 2,667*150/100. In line with the results of purity, CWFCM shows very satisfactory results, with fewer error rates in all datasets outperforming IFWFCM or FCM.

\subsection{Statistical Analysis}

To compare several methods on multiple datasets, this study uses a non-parametric statistical test, i.e. the Friedman test to decide differences in clustering problems. To get comprehensive and convincing results, the statistical analysis process uses the help of XLSTAT software. Table 11. shows the results of Friedman’s test for the evaluation method of error rates, from the table can explain that Friedman's test produces a p-value of 0.0035, smaller than the alpha value of 0.05, concluded for this condition is there are significant differences between methods.

To find out a comparison of which methods have significant differences, we use Nemenyi's post-hoc test so that the difference between method is clearly seen.A significant difference is if the p-values between the two methods are smaller than the 0.05 alpha value. In Table 12. it can be seen, the p-values between CWFCM and FCM are 0.0090 < 0.05 and the p-values between CWFCM and IFWFCM are 0.0483 <0.05.It is clear that the proposed method CWFCM has significant differences both with FCM and IFWFCM.

\begin{table}[h!]
	\caption{Friedman's test of error rates evaluation}
	\begin{center}
		\begin{tabular}{@{}ll@{}}
		\toprule
		Q (Observed value) & 11.2903 \\ \midrule
		Q (Critical value) & 5.9915 \\ \midrule
		DF & 2 \\ \midrule
		p-value (Two-tailed) & 0.0035 \\ \midrule
		alpha & 0.05 \\ \bottomrule
	\end{tabular}
	\end{center}
	\label{tab:table11}
\end{table}
\begin{table}[h!]
	\caption{P-values comparisons using nemenyi's procedure}
	\begin{center}
		\begin{tabular}{@{}|l|l|l|l|@{}}
		\toprule
		& FCM & IFWFCM & CWFCM \\ \midrule
		FCM & 1 & 0.8259 & \textbf{0.0090} \\ \midrule
		IFWFCM & 0.8259 & 1 & \textbf{0.0483} \\ \midrule
		CWFCM & \textbf{0.0090} & \textbf{0.0483} & 1 \\ \bottomrule
	\end{tabular}
	\end{center}
	\label{tab:table12}
\end{table}

In line with state-of-the-art of this study \cite{jafar2012study,das2013pattern,liao2013new}, that other distance metric is possible for using as distance metric in the FCM algorithm to produce better clustering. Evident in this study, that canberra distance has a good level of stability to decide the best distance between the data points and the cluster centers. This study also confirms the findings of previous research, that each feature will have a different weight in determining its class \cite{su2010new,hung2008bootstrapping,xing2014further}, it is proven by adding feature weights using the VMR, proposed method shows a very satisfying performance even in noisy datasets, multidimensional feature and have a lot of classes.

\section{Conclusions and feature works}

There are two major drawbacks that have been discovered by previous researchers, to improve performance of the FCM algorithm. In the terms of efficiency, random initialization is the first drawback, in terms of the accuracy of grouping data euclidean distance becomes the second drawback because it treats all features with the same weight. In this study, we propose a novel method that we called CWFCM, we used fuzziness of sum square root from each feature to initialize partition matrix and replaced the initialization process randomly. To overcome the second drawback, we propose by adding VMR to get feature weights that match the characteristics of each datasets and change the euclidean distance by canberra distance to get the optimum distance between data point with each cluster center. We have compared several distance measurements and feature weights that are often used in the FCM algorithm, on average the proposed method outperform even with the added noise up to 30$\%$.In addition, we have also compared the proposed method with the earlier method using various measuring instruments, and the results of proposed method outperform both in terms of computational time and accuracy in grouping data. The results of statistical tests using non-parametric statistical tests also show that the proposed method has a significant difference compared to the earlier method.

Clustering is the unsupervised learning that can predict the number of classes based on observation methods from the characteristics of the datasets. Furthermore its will be interesting to add an internal validation model that evaluated in the terms of automatically determining the number of clusters in the FCM algorithm.

\bibliographystyle{elsarticle-num}
\bibliography{mybibfile}

\end{document}